%% file: aaai22.tex
\def\year{2022}\relax
\documentclass[letterpaper]{article} 
\usepackage{aaai22}  
\usepackage{times}  
\usepackage{helvet}  
\usepackage{courier}  
\usepackage[hyphens]{url}  
\usepackage{graphicx} 
\usepackage{natbib}  
\usepackage{caption} 
\DeclareCaptionStyle{ruled}{labelfont=normalfont,labelsep=colon,strut=off} 
\usepackage{algorithm}
\usepackage{algorithmic}

\usepackage{newfloat}
\usepackage{listings}
\floatstyle{ruled}
\newfloat{listing}{tb}{lst}{}
\floatname{listing}{Listing}
\title{On the Efficacy of Small Self-Supervised Contrastive Models \\without Distillation Signals}
\author{
    {\large Haizhou Shi\textsuperscript{\rm 2}\thanks{This work was done when Haizhou Shi and Wenjie Zhu were interning at OPPO Research Institute, Shanghai.},
    Youcai Zhang\textsuperscript{\rm 1},
    Siliang Tang\textsuperscript{\rm 2}\thanks{Corresponding author.},
    Wenjie Zhu\textsuperscript{\rm{3}*},
    Yaqian Li\textsuperscript{\rm 1},
    Yandong Guo\textsuperscript{\rm 1},
    Yueting Zhuang\textsuperscript{\rm 2}
    }
    
}
\affiliations{

    \textsuperscript{\rm 1} OPPO Research Institute,
    \textsuperscript{\rm 2} Zhejiang University,
    \textsuperscript{\rm 3} New York University \\
    \{zhangyoucai, liyaqian, guoyandong\}@oppo.com, 
    \{shihaizhou, siliang, yzhuang\}@zju.edu.cn, 
    wz2140@nyu.edu
}

\usepackage{bibentry}
\input{commands}

\begin{document}

\maketitle

\input{sections/abstract}
\input{sections/intro}
\input{sections/related}

\input{sections/diff_big_small}
\input{sections/explore}
\input{sections/main_experiments}
\input{sections/conclusion}
\input{sections/acknowledgement}

\bibliography{aaai22}
\end{document}

%% file: commands.tex
\usepackage{xspace}

\newcommand{\ressuper}{ResNet152\xspace}
\newcommand{\reslarge}{ResNet50\xspace}
\newcommand{\ressmall}{ResNet18\xspace}
\newcommand{\vit}{{ViT}\xspace}
\newcommand{\moblarge}{{MobileNetV3\_large}\xspace}
\newcommand{\mobsmall}{{MobileNetV3\_small}\xspace}
\newcommand{\eff}{{EfficientNet\_B0}\xspace}
\newcommand{\deit}{{DeiT\_Tiny}\xspace}

\newcommand{\al}{{alignment}\xspace}
\newcommand{\uni}{{uniformity}\xspace}

\newcommand{\smaller}[1]{\fontsize{7pt}{1em}\selectfont{#1}}

\newlength\savewidth\newcommand\shline{\noalign{\global\savewidth\arrayrulewidth
  \global\arrayrulewidth 1pt}\hline\noalign{\global\arrayrulewidth\savewidth}}
\newcommand{\tablestyle}[2]{\setlength{\tabcolsep}{#1}\renewcommand{\arraystretch}{#2}\centering\footnotesize}

\usepackage{multirow}
\usepackage{subfigure}
\usepackage{pifont}
\newcommand{\cmark}{\ding{51}}%
\newcommand{\xmark}{\ding{55}}%
\newcommand{\rmark}{\ding{221}}%

\usepackage{amsmath,amsfonts,amssymb,bm}

\def\Lalign{{\mathcal L_{\textnormal{align}}}}
\def\Lialign{{\mathcal L_{\textnormal{intra-align}}}}
\def\Luniform{{\mathcal L_{\textnormal{uniform}}}}

\def\eqref#1{equation~\ref{#1}}

\def\1{\bm{1}}

\DeclareMathAlphabet{\mathsfit}{\encodingdefault}{\sfdefault}{m}{sl}
\SetMathAlphabet{\mathsfit}{bold}{\encodingdefault}{\sfdefault}{bx}{n}

\def\gL{{\mathcal{L}}}

\def\gP{{\mathcal{P}}}

\def\gX{{\mathcal{X}}}

\newcommand{\E}{\mathbb{E}}



%% file: sections/abstract.tex
\begin{abstract}
It is a consensus that small models perform quite poorly under the paradigm of self-supervised contrastive learning. Existing methods usually adopt a large off-the-shelf model to transfer knowledge to the small one via distillation. Despite their effectiveness, distillation-based methods may not be suitable for some resource-restricted scenarios due to the huge computational expenses of deploying a large model. In this paper, we study the issue of training self-supervised small models without distillation signals. We first evaluate the representation spaces of the small models and make two non-negligible observations: (i) the small models can complete the pretext task without overfitting despite their limited capacity and (ii) they universally suffer the problem of over clustering. Then we verify multiple assumptions that are considered to alleviate the over-clustering phenomenon. Finally, we combine the validated techniques and improve the baseline performances of five small architectures with considerable margins, which indicates that training small self-supervised contrastive models is feasible even without distillation signals. The code is available at \textit{https://github.com/WOWNICE/ssl-small}.

\end{abstract}

%% file: sections/intro.tex
\section{Introduction}
Recently, the development of self-supervised contrastive learning has empowered the models to learn a good representation space without the guidance of labels. Among them, the large models, e.g., \reslarge, \ressuper, \vit, have achieved comparable results as the supervised learning methods~\cite{chen2020simple, chen2020big, he2020momentum, chen2021empirical}. 

The small models, however, could not gain such good performance under the same training paradigm. In supervised learning, a small model is outperformed by its large counterpart by 9\% in terms of accuracy~(\mobsmall's 67.7\% v.s \reslarge's 76.1\%). However, in self-supervised learning, the gap between the small and the large is dramatic~(\mobsmall's 26.8\% v.s \reslarge's 67.5\%). 
\begin{figure}[t]
  \centering
  \includegraphics[width=.46\textwidth]{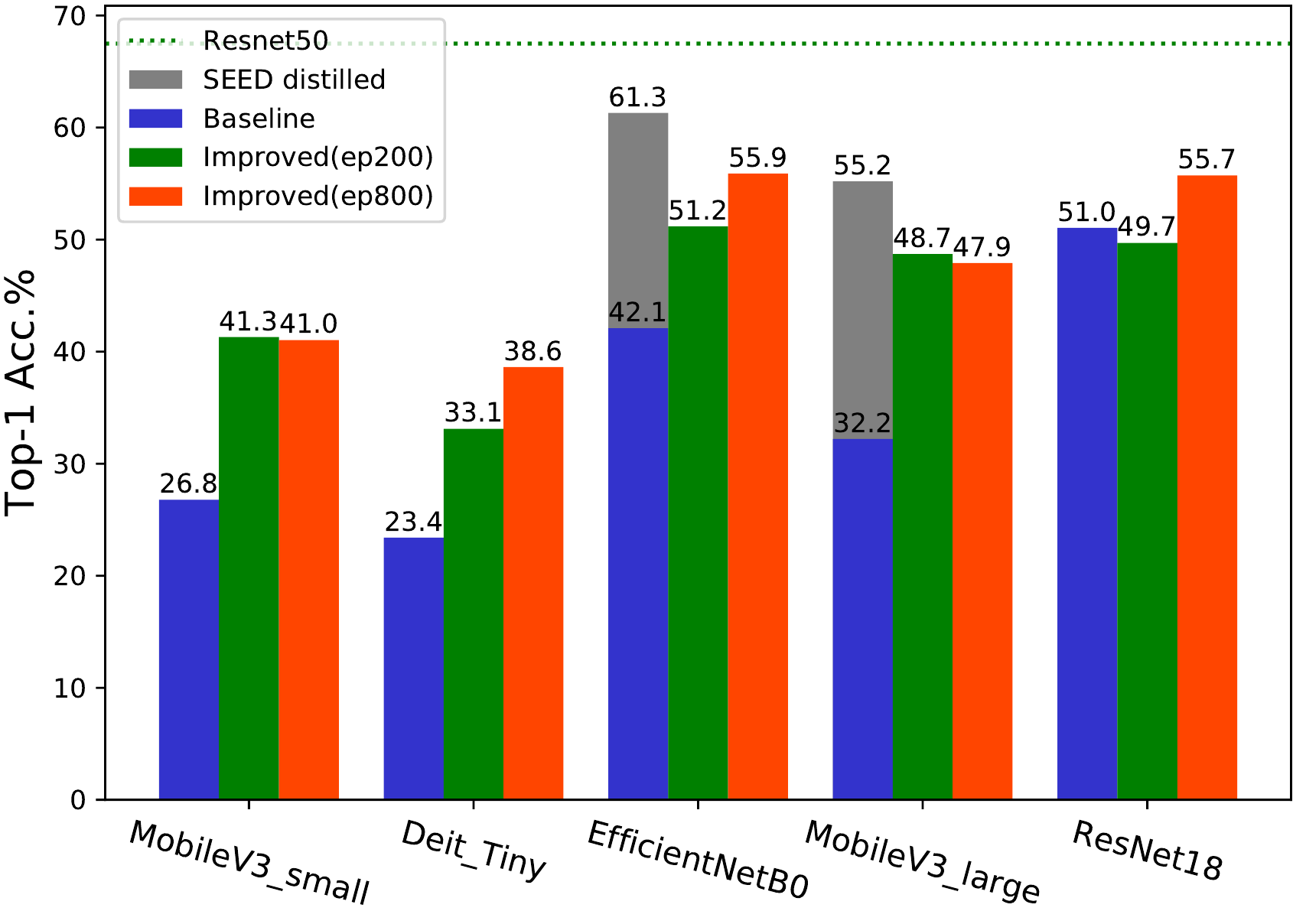}
  \caption{Baseline and improved performance of five popular small models. 
  The gray boxes represent the SEED distillation results with ResNet50 as teacher network.
  }
  \label{fig:overall}
\end{figure}
One simple and widely accepted assumption explaining small models' failure is proposed in many concurrent works~\cite{fang2021seed, xu2021bag, gu2021simple}. It argues that the pretext instance discrimination task is too challenging for the small models due to their limited capacity. Based on this assumption, the most popular framework of training self-supervised small models utilizes the technique of knowledge distillation~\cite{hinton2015distilling}, where the problem of training small self-supervised models boils down to two phases. It first trains a large learner in a self-supervised fashion and then trains the small learner to mimic its representation distribution~\cite{koohpayegani2020compress, fang2021seed, xu2021bag, gu2021simple, gao2021disco}. 

Although proven to be effective, this method is not applicable in some resource-restricted scenarios. For example, on mobile devices, due to the limited computational resource and data privacy, we cannot deploy a large model to guide the small model's learning or transfer the local data back to the server for distillation. Therefore, training small models with only contrastive learning signals is a challenge worth addressing.
Besides, we argue that distilling the knowledge from an already-known-working large model to a small model weakens the significance of the self-supervision setting since the large model, although trained without labels, is a strong supervision signal for the small one.

In this paper, we study the critical question of whether the small models can learn under the self-supervised contrastive signals without the guidance of the high-capacity teacher.
To this end, we first extensively study the properties of the representation spaces generated by the small and large model. We show that, contradictory to the aforementioned assumption that ``the small learners cannot complete the pretext task as well as the large learners'', small models can achieve comparable (or even better) pretext-task performance as the large ones. Despite this, the small learners seem to get stuck at the ``over-clustering'' position where the augmented views of the same sample are tightly clustered, whereas the samples of the same semantic class are not well clustered in the representation space.

Next, we study some of the common assumptions that are considered to solve the problem of over clustering, including higher temperature, fewer negative samples during training, more aggressive augmentation scheme, better weight initialization, and various projector architectures. We find that although some of the tricks work, there is no universal rationale behind the setting of these hyper-parameters, which showcases that the reason for the success of self-supervised contrastive learning still needs to be further understood. 
Nevertheless, we combine the validated tricks and finally improve the baseline performance of 5 often-used small learners by a large margin (on average $15+$ absolute linear evaluation accuracy points, measured on ImageNet1k~\cite{deng2009imagenet}), as shown in Fig.\ref{fig:overall}. Although there is still a performance gap between the self-supervised small models and their distillation-based counterparts, we show a huge potential of the research direction where the self-supervised small models are trained without distillation signals. Our contributions are summarized as follows:
\begin{itemize}
    \item To the best of our knowledge, our paper for the first time studies why the small models perform poorly under the current self-supervised contrastive learning framework. 
    \item We identify that small models can complete the pretext instance discrimination task as large ones. However, the representation spaces yielded by the small models universally suffer the problem of over clustering.
    \item We improve the baseline performance of 5 different small models by a large margin~(on average 15+ absolute points), showing the potential of training contrastive small models in the resource-restricted scenario. 
\end{itemize}

%% file: sections/related.tex
\section{Background and Related Work}
\subsection{Self-Supervised Contrastive Learning}
Contrastive learning adopts the multi-view hypothesis, which regards the augmented views of the data sample as positive and requires the model to distinguish them from other (negative) samples~\cite{arora2019theoretical}. The contrastive learning optimizes the following InfoNCE~\cite{oord2018representation} loss:
\begin{align}
    \gL_{\text{con}}(x_i) = -\log \frac{\exp(s_{i,i}/\tau)} {\exp(s_{i,i}/\tau) + \sum_{k\neq i}^{K} \exp(s_{i,k}/\tau)},
    \label{eq:infoNCE}
\end{align}
where $s_{i,i} = f(x_i)^\top f(x_i^+)$ and $(x_i, x_i^+)$ is the positive pair consisting of two randomly augmented views. $s_{i,j} = f(x_i)^\top f(x_j)$ is the similarity between two negative samples. The negative sample set $\{x_j\}^K_{j\neq i}$ is constructed by sampling for $K$ times independently from the data distribution. 

Based on the primary form of contrastive learning, multiple contrastive-based methods have been proposed to train the networks without supervision~\cite{wu2018unsupervised, oord2018representation, tian2019contrastive, chen2020simple, he2020momentum}. Many of them achieve the SOTA performance on the downstream linear classification task with the backbone network fixed~\cite{zhang2016colorful,oord2018representation,bachman2019learning}. However, little attention has been paid to training small models~\cite{howard2017mobilenets, tan2019efficientnet} solely under the contrastive learning framework, for its failure has been widely observed~\cite{koohpayegani2020compress, fang2021seed, gao2021disco, xu2021bag, gu2021simple}. In this paper, we want to fill in the void of training small models with and only with contrastive learning signals.

\subsection{Self-Supervised Small Models}
Currently, knowledge distillation~\cite{hinton2015distilling} becomes a widely acknowledged paradigm to solve the slow convergence and difficulty of optimization in self-supervised pretext task for small models~\cite{koohpayegani2020compress, fang2021seed, gao2021disco, xu2021bag, gu2021simple}. 
ComPress ~\cite{koohpayegani2020compress} and SEED~\cite{fang2021seed} distill the small models based on the similarity distributions among different instances randomly sampled from a dynamically maintained queue. 
DisCo~\cite{gao2021disco} removes the negative sample queue and straightforwardly distills the final embedding to transmit the teacher’s knowledge to a lightweight model. 
BINGO~\cite{xu2021bag} proposes a new self-supervised distillation method by aggregating bags of related instances to overcome the low generalization ability to highly related samples. 
SimDis~\cite{gu2021simple} establishes the online and offline distillation schemes and builds two strong baselines for the distillation-based training paradigm. 

We appreciate the efforts made by the previous researchers, and it is advisable to address the problem of training small self-supervised models in a divide-and-conquer way: [\emph{{Self-Supervised} {Small Model}}] = [\emph{{Self-Supervised} Large Model}] + [\emph{Supervised Distillation for {Small Model}}]. However, since the former and the latter tasks are both technically well-studied, we must point out that, by doing so, the existing works are somewhat evasive about the crucial problem ``why the small models benefit from the self-supervised method far less than the large models'' and ``how to improve them''. 
Furthermore, distillation-based methods are not applicable to some resource-restricted scenarios~({\it{e.g.}}, self-supervised learning on mobile devices) due to the huge computational expenses of deploying a large model, which is another core motivation of this work.

%% file: sections/diff_big_small.tex
\section{Representation Space Analysis}
To better understand why the small models perform poorly in self-supervised contrastive learning, we introduce several evaluation metrics that help us diagnose the problems. There are generally two types of metrics: the first being pretext-task-related metrics that do not require any human annotation and reflect how well the model performs on the instance discrimination, including (i) \textbf{alignment}, (ii) \textbf{uniformity}, and (iii) \textbf{instance discrimination accuracy}; the second category being downstream-task-related metrics that require semantic labels, including (i) \textbf{intra-class alignment}, (ii) \textbf{best-NN}, and (iii) \textbf{linear evaluation protocol}.


\subsubsection{Alignment.} 
The \al of the model is defined to measure the average squared distance of the samples' representations within a positive pair. It is one of the two training objectives of the contrastive loss under certain conditions~\cite{wang2020understanding}, and is the core of the multi-view hypothesis~\cite{tian2019contrastive}: 
\begin{align}
	\Lalign &\triangleq \E\left[ \|f(x_1)-f(x_2)\|_2^2 \right], \label{eq:align}
\end{align}
where $(x_1, x_2) \sim \gP_\text{pos}(x)$ denotes two augmented positive samples under the given data augmentation scheme.

\subsubsection{Uniformity.}
The \uni is defined to measure how uniform the representation distribution is in the hyper-spherical space~\cite{wang2020understanding}. It is the other objective the contrastive learning framework actively optimizes:
\begin{align}
	\Luniform &\triangleq \log \E\left[ \operatorname{exp}\left({-t\|f(x)-f(y)\|_2^2}\right) \right], \label{eq:uniform}
\end{align}
where $(x, y) \sim \gX^2$ denotes two data points sampled i.i.d from the data distribution $\gX$. The \al and the \uni of a representation space are essential to the model training; failing either one of them would cause the model to learn a non-generalizable representation space.

\subsubsection{Instance discrimination accuracy.}
The contrastive learning framework follows the instance discrimination pretext task~\cite{wu2018unsupervised,ye2019unsupervised}. It regards the augmented views of the same sample as of the same class and other samples as from different classes. Suppose there are $N$ samples in total, then the pretext instance discrimination task can be viewed as an $N$-way classification problem. Some methods measure this metric within a batch of samples during training~\cite{he2020momentum}, which could be problematic since the batch size hugely influences the value of the accuracy. This paper measures the above three metrics on a static pre-generated dataset from both training and validation set. We use the standard data augmentation scheme used in MoCoV2~\cite{he2020momentum} to create positive pairs. For a fair comparison, we sample 50 images per class for both the training set and the validation sets, making it a $50,000$-way classification task for the pre-trained models.

\subsubsection{Intra-class alignment.}
By \al, we measure whether augmented views of the same sample are mapped into a small and tight cluster. Similarly, we want to measure whether the samples of the same semantic classes are mapped to a close neighborhood in the representation space. Following the form of the \al term, we define the Intra-class alignment: 
\begin{align}
	\Lialign &\triangleq \E_{c, (x^{(c)}, y^{(c)})}\left[ \|f(x^{(c)})-f(y^{(c)})\|_2^2 \right], \label{eq:intra-class-align}
\end{align}
where $(x^{(c)}, y^{(c)}) \sim \gP_c$ denotes two samples that belong to the same semantic class $c$ sampled independently. Note our work does not originally propose this metric. The metric of \textit{tolerance} has been proposed in \citet{wang2021understanding}, where \textit{tolerance} $=1-\Lialign /2$ under the constraint of the hyper-spherical representation space. However, the definition of intra-class alignment complies more with our intuition since it has a similar tendency as \al. What's more, these two metrics have a more in-depth causal relationship: we often regard the optimization of intra-class alignment as the consequence of the optimization of alignment.

\subsubsection{Linear evaluation protocol.}
The linear evaluation protocol (also known as linear probing accuracy) is the standard metric that evaluates the linear separability of the representation space~\cite{zhang2016colorful,chen2020simple,he2020momentum}. It follows a basic assumption that the samples of different semantic labels should be easily separated in good representation spaces. In practice, we freeze the backbone network's parameter and train a simple linear classifier on top of it to measure the linear separability.

\subsubsection{Best-NN.}
In this paper, we extend the $k$-NN metric to the best-NN metric. Many works adopt $k$-NN as the indicator for the downstream task performance~\cite{fang2021seed,wu2018unsupervised} for it runs faster than the standard linear evaluation protocol and is deterministic once the hyper-parameter $k$ is determined. However, different methods may generate different types of representation space favoring different $k$ and cause unfair comparison. To solve this problem, we propose to use the best-NN metric, which picks the best $k$-NN accuracy out of the range $\{1, 3, \cdots, K\}$. In our setting, $K=101$:
\begin{align}
    \operatorname{best-NN} = \max_{k\in \{1, 3, \cdots,K\}}\left\{ k\operatorname{-NN} \right\}
\end{align}

\subsection{Differences Between Large and Small Models}
\begin{table}[t]
\begin{center}
\tablestyle{2.4pt}{1.25}
\begin{tabular}{c|c|ccccc}
 & res-50 & res-18 & mob-l & mob-s & eff-b0 & deit-t \\
\shline
\#params (M)            & 25 & 11 & 5.5 & 2.5 & 5.3 & 5 \\
\hline
training time (h) & {42.4} & {40.9} & {40.0} & {38.9} & {40.0} & {39.8}\\
\end{tabular}
\end{center}
\caption{Parameter number and the training time of various models. The training times are evaluated on a single 8-card V100 GPU server for 200 epochs of training.} 
\label{tab:time}
\end{table}

\begin{figure}[t]
  \centering
  \includegraphics[width=.44\textwidth]{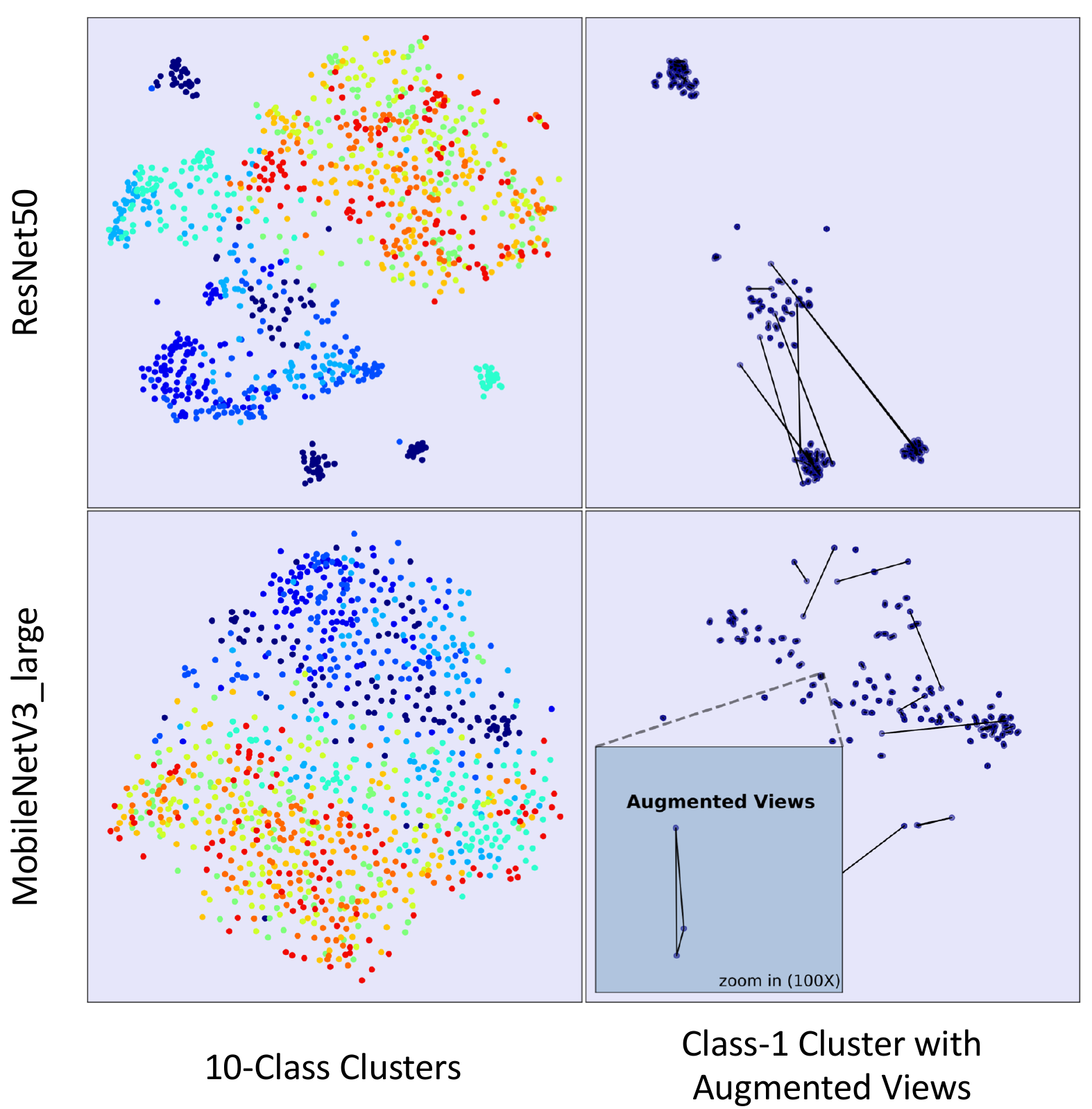} 
  \caption{t-SNE representation space visualization of the large and small model under the same setting, evaluated on 10 classes of ImageNet. \textbf{Top}: \reslarge; \textbf{Bottom}: \moblarge. We connect the original data point and its two augmented views as a triangle as shown in the bottom-right zoom-in window, which indicates the small model's representations are over-clustered. }
  \label{fig:tsne}
\end{figure}

We base our research on the MoCoV2 algorithm~\cite{chen2020improved} since it's computationally efficient and stable. To better utilize the computational resource, we set the batch size as 1024, and the learning rate as 0.06. The rest of the hyper-parameters are kept the same as the original MoCoV2 paper. We run the baseline experiments on five small models, including \ressmall~\cite{he2016deep} dubbed as res-18, \moblarge dubbed as mob-l, \mobsmall~\cite{howard2019searching} dubbed as mob-s, \eff~\cite{tan2019efficientnet} dubbed as eff-b0, and one small vision transformer model \deit~\citet{touvron2021training} dubbed as deit-t. For \deit, we set the head number of the transformer layer to 6 instead of 3 in the original work~\cite{dosovitskiy2020image,touvron2021training}. During training, we fix the linear mapping layer as suggested in \citet{chen2021empirical}. 
We train all the models on a single 8-card V100 GPU server, which well satisfies the computational requirement in all cases. We summarize the model parameter number~(\#params) and their training time as in Tab.\ref{tab:time}. Then we compare all the small models' behavior with the \reslarge large model (dubbed as res-50). All the metrics are evaluated on the penultimate output of the networks~(refer to Tab.\ref{tab:train-val}) and the ImageNet1k dataset~\cite{deng2009imagenet}. To yield a more intuitive understanding, we further visualize the representation distributions using the t-SNE~\cite{van2008visualizing} algorithm~(refer to Fig.\ref{fig:tsne}). We list our observations and corresponding conclusions as follows. \\

\begin{table}[t]
\begin{center}
\tablestyle{2pt}{1.25}
\begin{tabular}{c|cc|cc|cc|cc}
 &\multicolumn{2}{c|}{\includegraphics[scale=0.35]{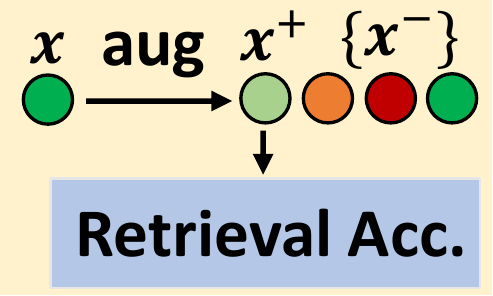}}  &\multicolumn{2}{c|}{\includegraphics[scale=0.35]{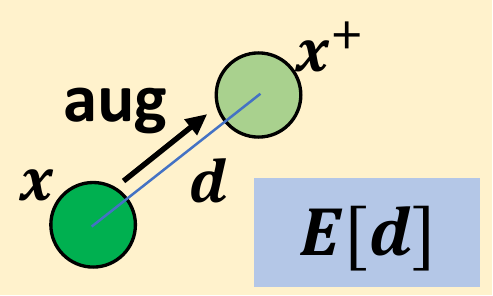}} &\multicolumn{2}{c|}{\includegraphics[scale=0.35]{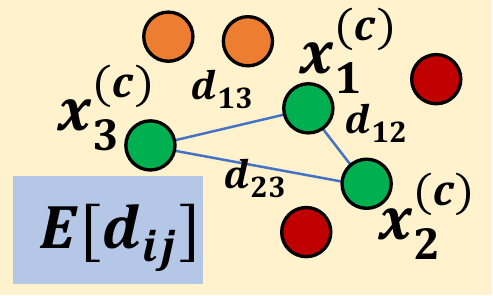}} &\multicolumn{2}{c}{\includegraphics[scale=0.35]{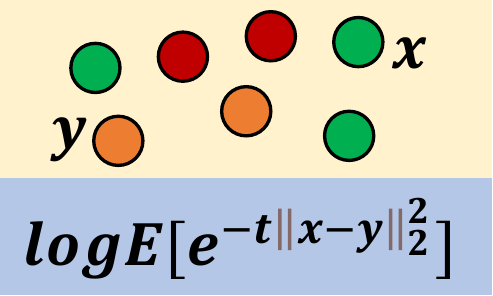}}\\

\multicolumn{1}{l}{} &\multicolumn{2}{|c|}{\smaller{inst-disc}}  &\multicolumn{2}{c|}{\smaller{align}} &\multicolumn{2}{c|}{\smaller{intra-class align}} &\multicolumn{2}{c}{\smaller{uniformity}}\\
&train  &val  &train &{val}   &{train}  &{val}    &{train}  &{val}\\
\shline
res-50	&83.4 &85.4 &0.25 &0.25 &0.91 &0.90 &-2.89 &-2.85\\
res-18	&81.2 &82.8 &0.24 &0.24 &0.95 &0.93 &-2.77 &-2.74\\
mob-l	&77.2 &78.8 &0.25 &0.25  &1.24 &1.23 &-3.29 &-3.28\\
mob-s	&86.2 &88.5 &0.25 &0.25 &1.32 &1.30 &-3.40 &-3.32\\
eff-b0	&81.0 &82.1 &0.26 &0.26 &1.29 &1.27 &-3.48 &-3.51\\
deit-t	&64.7 &68.1 &0.46 &0.46 &1.34 &1.34 &-3.06 &-3.09
\end{tabular}
\end{center}
\caption{Self-supervised baseline models trained by MoCoV2. Different colors demonstrate different semantic classes. The downstream-task performance of the baseline models are shown in Tab.\ref{tab:main}. 
}
\label{tab:train-val}
\end{table}

\noindent\textit{OBSERVATION. The large and small models perform similarly on the pretext instance discrimination task.} 

The rationale behind self-supervised learning is that completing a human-designed challenging pretext task forces the model to learn the ability to extract generalizable features from the data. In the instance discrimination setting, models are required to do a $N$-way classification task where $N \gg C$, $N$ is the number of the samples, and $C$ is the number of the semantic classes. Therefore, when given the fact that the small models perform way worse than the large models, it’s natural to conjecture that the pretext contrastive learning task might be too difficult for them. However, we are surprised to find that under the same training setting, the small learners, despite the architecture differences, can complete the instance discrimination task as well as the large model, in some cases (Tab.\ref{tab:train-val}, \textbf{mob-s}' 86.2\% v.s \textbf{r-50}'s 83.4\%) even better than it. This phenomenon concludes that the pretext contrastive learning task is an easy task, (except for \deit). It also leads to a weird conclusion that accomplishing the pretext contrastive learning task doesn't have an absolute positive correlation with the model's downstream generalization ability. If this conclusion holds, then more critical questions about contrastive learning should be posed: (i) what else is optimized during self-supervised contrastive learning, and (ii) what is critical to the generalization other than the instance-discriminative ability?\\ 

\noindent\textit{OBSERVATION. For the small models, there is no overfitting problem when trained on the pretext task.} 

This conclusion is supported by the fact that each model's metrics have no significant difference on both the training and validation sets. According to recent studies, one of the most significant differences between large networks and small networks is their generalization ability. The phenomenon of ``double descent'' has been observed by the deep learning researchers~\cite{nakkiran2019deep}. The large networks can reduce the generalization error when the number of the model parameters increases to a certain extent. However, the concept of the generalization error, which measures the difference between the training and the testing error, cannot explain the small self-supervised models' poor performance on the downstream tasks. Based on this observation, we will only measure the models on the validation set in the following sections. \\

\noindent\textit{OBSERVATION. The small models' representation spaces are universally over-clustered compared to the large model.}

According to Tab.\ref{tab:train-val}, we can see that the \al of the small models is almost the same as the large model. However, the intra-class alignment is significantly higher than the large model. It showcases that the small models address the pretext-instance discrimination problem by trivially mapping augmented views of the same sample into a tight cluster without effectively clustering the samples of the same semantic label.
This conclusion is supported by Fig.\ref{fig:tsne} as well.
By comparing the subfigures in the left column, we can see that \reslarge achieves a representation space where same-class samples are well clustered. While \moblarge's representations are loosely scattered, the decision boundaries of which are not clear. 
As defined in \citet{wang2021solving}, this could be caused by either under-clustering (positive samples are not properly aligned) or over-clustering (positive samples are well aligned, but not leading to proper class-level clustering). To identify the cause, we further visualize the representation with their augmented views in the right column. The triangles are formed by connecting the samples and their positive pair. The zoom-in window~(100X) shows that \moblarge can perfectly align the positive pairs of data, which proves the existence of over-clustering problem. 

%% file: sections/explore.tex
\section{Solve Over Clustering with Simple Assumptions}
Based on the above observations, we approach the over-clustering problem by applying simple tricks that seem to have a direct influence in this section. The assumptions are presented in the order of their importance: the temperature has the greatest influence and therefore we study it and fix it for the remaining assumptions; the negative samples and data augmentation correspond to the fundamental setting of negative/positive construction for contrastive learning; finally we deal with the network structure and weight initialization. At the end of the statement of each assumption, we use \textit{[\xmark]} \& \textit{[\cmark]} to pre-indicate the correctness based on its following experiments.

\begin{figure}[t]
  \centering
  \includegraphics[width=.46\textwidth]{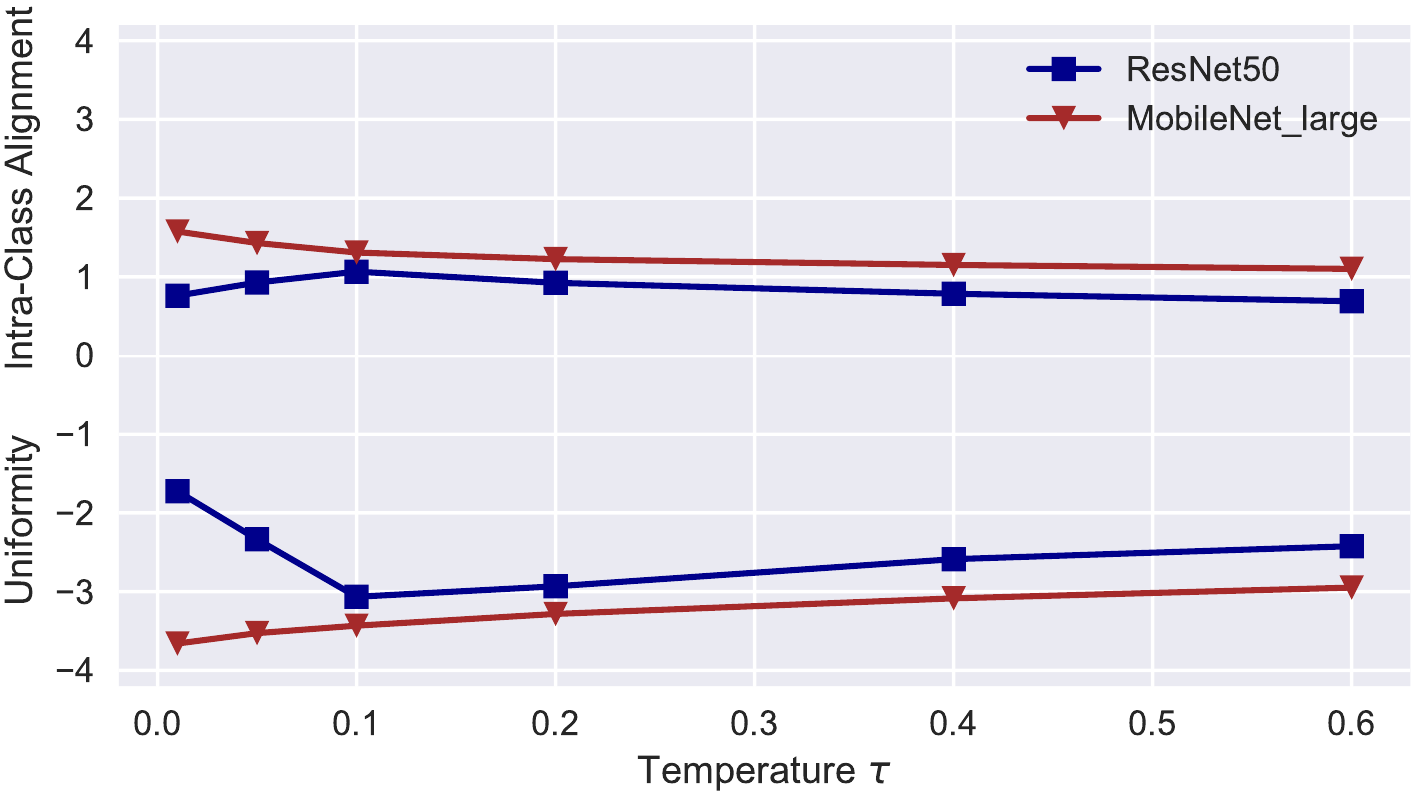} 
  \caption{Intra-class alignment and uniformity yielded by different temperature $\tau$. }
  \label{fig:uni-align}
\end{figure}

\subsection{Temperature}
\textit{ASSUMPTION. The temperature trades off the intra-class alignment and the uniformity. Higher temperatures can alleviate the problem of over-clustering. [\xmark]}

\citet{wang2021solving} partially demonstrates how the temperature influences the contrastive model's behavior. When the temperature is infinitely large or infinitely close to zero, the original contrastive loss will degrade to two simple forms: 
\begin{align}
    \gL_{\text{simple}}(x_i) &= -s_{i,i} + \lambda \sum_{k\neq i}s_{i,k}, \quad (\tau \rightarrow +\infty)\\
    \gL_{\text{triplet}}(x_i) &= \max\left[ s_{\max}-s_{i,i}, 0 \right], \quad (\tau \rightarrow 0^+)
\end{align}
where in the first case, the model pays equal attention to the negative samples. In contrast, the model ignores all the negative samples but the one with the maximum similarity in the second case. Based on the analysis, lower temperature causes the model to pay more attention to the negative samples that are close to the current data point; higher temperature causes the model to pay more uniform attention to all the negative samples. Their experiments show that a higher temperature can improve the intra-class alignment by sacrificing the overall uniformity of the representation space, which means alleviating the problem of over-clustering. To validate this idea, we run experiments on different temperature setups on \moblarge and \reslarge. The results are shown in Tab.\ref{tab:tau}. 

\begin{table}[t]
\begin{center}
\tablestyle{6pt}{1.25}
\begin{tabular}{c|c|cccccc}
& & \multicolumn{6}{c}{Temperature $\tau$} \\
& & 0.01 & 0.05    & 0.1    & 0.2   & 0.4   & 0.6   \\
\shline
\multirow{3}{*}{mob-l} & linear  &24.3  &33.5 &\textbf{36.7} &31.3 &21.4 &15.9\\
& NN   &25.5  &32.8 &\textbf{32.9} &28.0 &21.2 &17.3\\
& inst-disc &91.6  &\textbf{93.3} &90.8 &78.8 &63.2 &50.1\\
\hline
\multirow{3}{*}{r-50} & linear  &43.9  &60.6 &63.8 &\textbf{67.5} &64.3 &61.8\\
& NN    &31.3  &41.0 &47.8 &\textbf{50.5} &47.4 &43.9\\
& inst-disc &90.0  &89.2 &\textbf{91.3} &84.0 &80.1 &76.1
\end{tabular}
\end{center}
\caption{The performance of the large and small models trained by different temperatures. 
``\textbf{linear}'': top-1 accuracy of the linear evaluation protocol; 
``\textbf{NN}'': top-1 accuracy of the Best-NN;
``\textbf{inst-disc}'': top-1 accuracy of the pretext instance discrimination task.}
\label{tab:tau}
\end{table}

There are two main takeaways. First, the temperature interacts with the contrastive learning methods in a more complicated way than we expected. Increasing the temperature does lead to a less uniform representation space with better intra-class alignment for the small model. However, for the large model, this principle does not fully apply: when the temperature is in the range of $(10^{-2}, 10^{-1})$, increasing the temperature causes a more uniform representation space, which is against the previously accepted understanding~\cite{wang2021understanding}. Secondly, for the small learner, although it suffers from the problem of over-clustering, further trading-off the intra-class alignment for better uniformity leads to better downstream-task performance. It indicates that generally, we want to keep the temperature lower than the large model for the small model.  
The reason for this phenomenon might be due to the slow convergence of the small model. As shown in \citep{wang2021understanding}, the gradient w.r.t the positive pair and the negative pairs are proportional to the reciprocal of the temperature $1/\tau$. We conjecture that lowering the temperature might have a similar effect of increasing the learning rate and thus help the model to converge faster. We leave the in-depth study on this subject for future. We set the temperature $\tau=0.05$ based on its best-NN and inst-disc performance for the rest of this section. 



\subsection{Negative Sample Size}
\textit{ASSUMPTION. The over-clustering problem is caused by the instance discrimination setting. False-negative samples make it hard for the small models to cluster data of the same semantic class. [\xmark]}

A lot of work have focused on analyzing the influence of the current negative sampling strategies ~\cite{wu2020conditional,wang2021solving,wang2021understanding,huynh2020boosting}. According to ~\cite{wang2021solving}, reducing the negative sample size would alleviate the problem of over clustering since it reduces the probability of colliding into false-negative samples in the mini-batch of data. Although it seems not to be a problem for the large model~(larger the negative sample size, better the performance~\cite{chen2020simple}), we cannot exclude the possibility that the small models might be more prone to the problem of false-negative samples. We try different negative sample sizes in this section and find that, contrary to the large model's benefiting from a larger number of the negative sample~\cite{he2020momentum}, the small model's performance fluctuates with the number of negative samples, which counters the proposed assumption.   
\begin{table}[h]
\begin{center}
\tablestyle{6pt}{1.25}
\begin{tabular}{c|cccc}

& 65536 & 16384 & 4096 & 1024\\
\shline
linear & 33.5 & \textbf{34.2} & 33.0 & 33.5\\
NN     & 32.8 & \textbf{33.3} & 32.6 & 33.1 \\
\end{tabular}
\end{center}
\caption{Results of different negative sample sizes.}
\label{tab:neg}
\end{table}

\subsection{Data Augmentation}
\textit{ASSUMPTION. The current data augmentation scheme is not challenging enough. It cannot drive the small model to learn from the self-supervised signal continually. [\cmark]}

Following \citet{tian2020makes}, we mainly consider the influence of the color distortion. As shown in Tab.\ref{tab:aug}, we can see that increasing the strength of the data augmentation can help the small model to achieve better linear separability. However, there is also a sweet spot in selecting the augmentation scheme. If the augmentation is too strong, it may violate the basic ``multi-view'' assumption, and the model will learn the hand-crafted noises. If the augmentation is too weak, the model will easily align the positive samples without learning high-level visual patterns. 

\begin{table}[t]
\begin{center}
\tablestyle{5pt}{1.25}
\begin{tabular}{cccccc|cc}
\multicolumn{4}{c}{CJ} & 
\multirow{2}{*}{GS} & 
\multirow{2}{*}{GB} & 
\multirow{2}{*}{linear} & 
\multirow{2}{*}{NN} \\ 
\multicolumn{1}{c}{\smaller{brightness}} & 
\multicolumn{1}{c}{\smaller{contrast}} & 
\multicolumn{1}{c}{\smaller{saturation}} & 
\multicolumn{1}{c}{\smaller{hue}} &  &  &  &  \\
\shline
\multicolumn{1}{c}{0.4} &\multicolumn{1}{c}{0.4} & \multicolumn{1}{c}{0.4} & \multicolumn{1}{c}{0.1} &  0.2 & 0.5 & 33.5 & 32.8\\
\multicolumn{1}{c}{0.4} &\multicolumn{1}{c}{0.4} & \multicolumn{1}{c}{0.4} & \multicolumn{1}{c}{0.2} & 0.2 & 0.5 & 34.8 & 34.8\\
\multicolumn{1}{c}{0.8} &\multicolumn{1}{c}{0.8} & \multicolumn{1}{c}{0.8} & \multicolumn{1}{c}{0.4} & 0.5 & 0.5 & \textbf{36.6} & \textbf{35.7}\\
\multicolumn{1}{c}{0.8} &\multicolumn{1}{c}{0.8} & \multicolumn{1}{c}{0.8} & \multicolumn{1}{c}{0.4}  & 0.5 & 1 & 35.2 & 35.0\\
\end{tabular}
\end{center}
\caption{Results of different data augmentation. 
``\textbf{CJ}'': ColorJitter;
``\textbf{GS}'': Grayscale;
``\textbf{GB}'': GaussianBlur.
}

\label{tab:aug}
\end{table}

For the small model, we conjecture that due to its limited capacity, although it can perform comparably on the pretext instance discrimination task as the large model, it might achieve this goal by taking a different learning path. It has been validated that if we do not apply the color distortion when creating the positive pairs, the model can fit the training objective quickly by taking the shortcut of color abbreviation~\cite{chen2020simple, tian2020makes}. Therefore we argue that the small models might be more prone to the shortcut existing in contrastive learning. Thus adding more randomness into the data augmentation is conducive to alleviating this problem.

\subsection{Weight Initialization}
\textit{ASSUMPTION. The small models are more prone to getting stuck at the local minimum during self-supervised contrastive training. [\cmark]}

Usually, a good initialization point helps the model converge to a better minimum point. To validate this idea, we first train the small model under the supervision of a pretrained large model using the SEED loss~\cite{fang2021seed} for certain epochs. Then we load its weights and train the small model purely under the contrastive learning framework~(same hyperparameter setting, no distillation signal). We benchmarked three sets of weights, which are trained under SEED for 2, 10, and 100 epochs, respectively. 

\begin{table}[h]
\begin{center}
\tablestyle{10pt}{1.25}
\begin{tabular}{c|cccc}
 & 0 & 2 & 10 & 100 \\
\shline
NN-init & -    & 30.2 & 43.1 & 47.5 \\
\hline
NN      & 32.8 & 35.0 & 37.8 & \textbf{39.3} \\
linear  & 33.5 & 35.4 & 39.4 & \textbf{42.0}
\end{tabular}
\end{center}
\caption{Results of initialization using weights trained by SEED. ``\textbf{NN-init}'': the best-NN performance after initialized with the SEED weights, before SSL training.} 
\label{tab:init}
\end{table}

As shown in Tab.\ref{tab:init}, the model initialized by a better SEED checkpoint yields better downstream performance. When the weights are trained by SEED for 10 and 100 epochs, the final downstream accuracy doesn't surpass the accuracy before SSL training. It also demonstrates that the convergence point of the distillation-based methods and the SSL methods are different. One may argue that it's pointless to reload the small model from the distilled checkpoint and train it on the pure SSL signal. However, in reality, the small models are often deployed on the edge devices whose data distributions are much different from the central server. We can first train the small models using distillation loss and then deploy them onto edge devices if the distilled weights can help the small model to converge faster.

\begin{table*}[t]
\begin{center}
\tablestyle{3.1pt}{1.25}
\begin{tabular}{c|cc|cc|cc|cc|cc}
 & 
 \multicolumn{2}{c|}{\smaller{MobileNetV3\_large}} &  
 \multicolumn{2}{c|}{\smaller{MobileNetV3\_small}} & 
 \multicolumn{2}{c|}{\smaller{EfficientNet\_B0}} & 
 \multicolumn{2}{c|}{\smaller{ResNet18}} & 
 \multicolumn{2}{c}{\smaller{DeiT\_Tiny}}
 \\
 &
\multicolumn{1}{c}{linear} &
\multicolumn{1}{c|}{NN} &
\multicolumn{1}{c}{linear} &
\multicolumn{1}{c|}{NN} &
\multicolumn{1}{c}{linear} &
\multicolumn{1}{c|}{NN} &
\multicolumn{1}{c}{linear} &
\multicolumn{1}{c|}{NN} &
\multicolumn{1}{c}{linear} &
\multicolumn{1}{c}{NN} 
\\

\shline
baseline (200)
    & 32.2 & 28.0
    & 26.8 & 23.5 
    & 42.1 & 31.5
    & 51.1 & 39.3 
    & 23.4 & 17.4 \\
\hline
improved  (200)
    & \textbf{48.7} {\smaller{(+16.5)}} 
        & 39.1 {\smaller{(+11.5)}}
    & \textbf{41.3} {\smaller{(+14.5)}} 
        & 32.8 {\smaller{(+9.3)}} 
    & 51.2 {\smaller{(+9.1)}} 
        & 41.3 {\smaller{(+9.8)}}
    & 49.7{\smaller{(-1.3)}} 
        & 39.6 {\smaller{(+0.3)}}  
    & 33.1 {\smaller{(+9.7)}} 
        & 30.5 {\smaller{(+13.1)}} \\
improved~(\textbf{800}) 
    & 47.9 {\smaller{(+15.7)}} 
        & \textbf{41.6} {\smaller{(+13.5)}}
    & 41.0 {\smaller{(+14.3)}} 
        & \textbf{35.6} {\smaller{(+12.0)}} 
    & \textbf{55.9} {\smaller{(+13.8)}} 
        & \textbf{44.2} {\smaller{(+12.7)}}
    & \textbf{55.7} {\smaller{(+4.7)}} 
        & \textbf{44.3} {\smaller{(+5.0)}}  
    & \textbf{38.6} {\smaller{(+15.2)}} 
        & \textbf{33.4} {\smaller{(+16.0)}} \\
\end{tabular}
\end{center}

\caption{Linear evaluation results of improved baseline performance of five different small self-supervised contrastive models, pretrained and evaluated on ImageNet. The number listed in the parenthesis indicates the the epochs of pretraining.}
\label{tab:main}

\end{table*}

\begin{table*}[t]

\begin{center}
\tablestyle{5.9pt}{1.25}
\begin{tabular}{c|ccc|ccc|ccc|ccc|ccc}
 &  \multicolumn{3}{c|}{\smaller{MobileNetV3\_large}}&  
 \multicolumn{3}{c|}{\smaller{MobileNetV3\_small}}& 
 \multicolumn{3}{c|}{\smaller{EfficientNet\_B0}}& 
 \multicolumn{3}{c|}{\smaller{ResNet18}}& \multicolumn{3}{c}{\smaller{DeiT\_Tiny}}
 \\
 &
\multicolumn{1}{c}{\smaller{C10}} &
\multicolumn{1}{c}{\smaller{C100}} &
\multicolumn{1}{c|}{\smaller{Cal101}} &
\multicolumn{1}{c}{\smaller{C10}} &
\multicolumn{1}{c}{\smaller{C100}} &
\multicolumn{1}{c|}{\smaller{Cal101}} &
\multicolumn{1}{c}{\smaller{C10}} &
\multicolumn{1}{c}{\smaller{C100}} &
\multicolumn{1}{c|}{\smaller{Cal101}} &
\multicolumn{1}{c}{\smaller{C10}} &
\multicolumn{1}{c}{\smaller{C100}} &
\multicolumn{1}{c|}{\smaller{Cal101}} &
\multicolumn{1}{c}{\smaller{C10}} &
\multicolumn{1}{c}{\smaller{C100}} &
\multicolumn{1}{c}{\smaller{Cal101}} \\

\shline
baseline (200)  & 71.8 & 42.4 & 72.9 & 70.0 & 40.4 & 70.0 & 72.0 & 43.2 & 77.2 & 81.5 & 54.0 & 81.2 & 67.2 & 40.3 & 62.1 \\
\hline
improved (200) & 77.9 & 51.4 & 84.0 & 75.5 & 48.5 & 81.0 & 78.3 & 51.6 & 86.0 & 79.9 & 52.2 & 85.1 & 76.8 & 50.0 & 78.5 \\
improved~(\textbf{800}) & \textbf{80.3} & \textbf{53.8} & \textbf{85.3} & \textbf{78.6} & \textbf{52.2} & \textbf{82.7} & \textbf{81.5} & \textbf{55.6} & \textbf{86.9} & \textbf{82.0} & \textbf{54.4} & \textbf{86.7} & \textbf{79.7} & \textbf{54.7} & \textbf{81.0}\\

\end{tabular}
\end{center}
\caption{Transfer learning results of the improved baselines across different architectures. 
``\textbf{C10}'': evaluated on CIFAR10 dataset;
``\textbf{C100}'': evaluated on CIFAR100 dataset;
``\textbf{Cal101}'': evaluated on Caltech101 dataset. 
All the models are pretrained and evaluated on ImageNet. The number listed in the parenthesis indicates the the epochs of pretraining..
}
\label{tab:transfer}
\end{table*}

While in this work, to make sure the comparison between the loaded model and the baseline model is fair, we only use the weights that have worse downstream-task performance than the baseline model as the initialization weights in both the main experiments and the ablation study.

\subsection{Projector Architecture}
\textit{ASSUMPTION. Deeper~\cite{fang2021seed} [\xmark] / Wider ~\cite{gao2021disco} [\cmark] / Dropout~\cite{gao2021simcse}[\xmark] MLP projector brings improvement to the (small) models. }

The rationale behind the deeper/wider projector is simple: adding more parameters will make small models somewhat closer to the large model (without considering the architecture difference). Apart from this, one theory~\cite{fang2021seed} states that the top layer of the network is more focused on addressing the pretext-task and loses some of the generalization ability. A deeper MLP projector would make the backbone network's representation farther away from the top layer and yield better downstream performance. A wider MLP obeys the Information Bottleneck~(IB) principle~\cite{gao2021disco} and thus make the small learner more capable of preserving the information, which might lead to better performance. As for the dropout layer in the projector, one can regard it as equivalent to a stronger data augmentation, creating a more challenging pretext task for the backbone network. We validate these assumptions one by one: 

\begin{table}[h]
\begin{center}
\tablestyle{9pt}{1.25}
\begin{tabular}{c|c|cc}
&  MLP projector structure & \multicolumn{1}{c}{linear} &  \multicolumn{1}{c}{NN} \\
\shline
baseline & 1280, 128 & 33.5 & 32.8 \\
\hline
\multirow{1}{*}{deeper} & 1280,1280,128 & 29.0 & 28.4 \\
\hline
\multirow{2}{*}{wider} & 2560, 128 & \textbf{33.9} & \textbf{33.2}\\
& 1280,256 & 32.8 & 31.7 \\
\hline
\multirow{1}{*}{dropout}& 1280, dropout(p=0.5), 128  & 28.9 & 26.7 \\
\end{tabular}
\end{center}
\caption{Results of different MLP projector architectures.} 

\label{tab:mlp}
\end{table}

In Tab.\ref{tab:mlp}, only the projector with the wider intermediate layer can improve the model with a small margin. One thing confusing about the dropout MLP is that it indeed creates a more challenging pretext task for the model and yields better alignment~(0.32\rmark 0.28), but worse intra-class alignment~(1.43\rmark 1.57). Usually, we assume the optimization of alignment causes the improvement of intra-class alignment since the former is the direct training objective in contrastive learning. However, the phenomenon produced by the dropout layer might challenge the common understanding and force us to rethink the efficacy of contrastive learning.

\subsection{Summary}
In summary, the assumptions that help the small model to achieve better downstream-task performance are listed as follows: (i) lower temperature, (ii) MLP projector with wider architecture, (iii) stronger augmentation scheme, and (iv) better weight initialization slightly trained by distillation-based methods. We will then in the next section combine the validated assumptions and apply them to 5 small models. 

%% file: sections/main_experiments.tex
\section{Improved Baselines for Small Self-Supervised Contrastive Models}
\label{sec:main}

\subsubsection{In-domain downstream classification.}
We first validate the summarized measures for five different small models in the downstream classification task. To showcase their effectiveness, we adopt the same training hyper-parameter set for all the small architectures without individually tuning them. We set temperature $\tau=0.1$, batch size $B=512$. learning rate $\eta=0.06$, and negative sample size $K=65536$. For wider MLP projector, we adaptively set the intermediate layer of the MLP as twice wide as its input layer for all the architectures. For augmentation scheme, we adopt the best color distortion scheme as the strengthened ``aug+''. We use the checkpoints that are trained by SEED for two epochs as the weight initialization for all the models except \deit since we observe that the small vision transformer has much slower convergence rate than other models even being trained by strong distillation signal; we train the \deit by SEED for 20 epochs and then use it as the weight initialization for SSL training~\cite{dosovitskiy2020image,touvron2021training}. We train all the models for 800 epochs with cosine decay, and evaluate them at epoch 200 and epoch 800. 

In Tab.\ref{tab:main}, all five models are improved by a large margin compared to the original baselines, which showcases the effectiveness and universality of our work. One thing to note here, training longer epochs (800) will surely improve the best-NN accuracy, while it's not the case for the linear evaluation protocol: the MobileNetV3 architectures will lose some of the linear separability to longer training.

\begin{table}[t]
\begin{center}
\tablestyle{3.4pt}{1.25}
\begin{tabular}{c|cccc|cc|c}
~  & \multicolumn{4}{c|}{\smaller{mocov2. pre-train}} & \multicolumn{2}{c|}{\smaller{ImageNet Acc.}} & \multicolumn{1}{c}{\smaller{Transfer Acc.}} \\

$\tau$ &
w-MLP &
aug+ &
init &
epochs & 
\multicolumn{1}{c}{linear} & 
\multicolumn{1}{c|}{NN} & 
cifar100  \\
\shline
 0.2 & & & & 200 & 32.2 & 28.0 & 42.4 \\
\hline
\multirow{5}{*}{0.05} & & & & 200 & 33.5 & 32.8 & 44.6 \\
& \cmark & & & 200 & 33.9 & 33.2 & 45.7\\
& & \cmark & & 200 & 36.6 & 35.7 & 47.0 \\
& & & \cmark & 200 & 35.4 & 35.0 & 48.3\\
& \cmark & \cmark & \cmark & 200 & \textbf{38.0} & \textbf{37.6} & \textbf{49.3} \\
\hline
\multirow{4}{*}{0.1} & & & & 200 & 36.7 & 32.9 & 46.4\\
& \cmark & \cmark &  & 200 &42.6 & 37.5 & 50.6\\
& \cmark & \cmark & \cmark & 200 & 46.0 & 40.3 & 51.8\\
& \cmark & \cmark & \cmark & \textbf{800} & \textbf{47.9} & \textbf{41.6} & \textbf{53.9}\\
\end{tabular}
\end{center}
\caption{Ablation study of \moblarge trained by MoCoV2.
``\textbf{w-MLP}'': with a wider MLP head;
``\textbf{aug+}'': with augmentation having stronger color distortion as in section~\cite{tian2020makes};
``\textbf{init}'': with better initialization trained by SEED for 2 epochs.
}
\label{tab:ablation}
\end{table}

\subsubsection{Transfer learning for classification.}
We benchmark the transferability of the backbone networks on CIFAR10, CIFAR100~\cite{krizhevsky2009learning}, and Caltech101~\cite{fei2004learning} image classification datasets. 
We evaluate the accuracy of the transfer learning in the same way as the linear evaluation protocol. To make the same set of hyper-parameters usable for different models, we follow the MoCoV2~\cite{chen2020improved} and apply normalization to the representations before they are put into the linear classifier. We first tune the hyper-parameters, including base learning rate and the learning schedule for the pretrained ResNet50 model to make sure it is comparable to the value reported in the SimCLR~\cite{chen2020simple}. Then we fix the hyper-parameters and apply them to all the small models. 

We apply bicubic interpolation when resizing the image to 224x224, random cropping of scale $[0.6, 1]$, and random horizontal flipping during training. During testing, we first resize the image to 256x256 and center-crop it to 224x224. For CIFAR10/CIFAR100, we set the batch size $B=256$, training epochs $E=60$, base learning rate $\eta=1.5$, and decrease the learning to one-tenth at epoch 40. For Caltech101, we set the base learning rate as 2 without scheduling. The results of transfer learning are reported in Tab.\ref{tab:transfer}, which reflects that the models trained by our improved setting have a more generalizable representation space.

\subsubsection{Ablation study.}
We present the ablation study to verify the effectiveness of all the working measures we find, as shown in Tab.\ref{tab:ablation}. We can see that our simple tricks can improve the baseline by a large margin on both downstream classification tasks and transfer learning in both temperature settings. Lower temperature, better initialization, and more aggressive data augmentation are the factors that have the strongest influence on improving the small models' representation quality.


%% file: sections/conclusion.tex
\section{Conclusion}
This work does not propose new technical contributions. Instead we provide an in-depth analysis on the behavior of the small self-supervised models with rigorous empirical verification: why do they fail and how can we improve them? Supported by empirical evidence, we point out a non-negligible fact that the small models can address the pretext instance discrimination task, and they do not overfit on its training data. However, they universally suffer the problem of over-clustering and therefore yield poor-quality representations. We then experiment with several assumptions that are supposed to solve this problem. Finally, we summarize and combine the practical measures, considerably improving the current baselines for five distinct small models. Our work shortens the gap between the small models and their large counterparts in self-supervised learning, highlighting that training small models without a high-capacity teacher model is a promising direction of research.

%% file: sections/acknowledgement.tex
\section{Acknowledgements}
This work has been supported in part by the National Key Research and Development Program of China (2018AAA0101900), Zhejiang NSF (LR21F020004), Chinese Knowledge Center of Engineering Science and Technology (CKCEST).